\documentclass{article}

\usepackage{amsmath,amsfonts,amssymb,times,graphicx,natbib,algorithm,algorithmic,hyperref}

\usepackage[accepted]{whi2018}
\usepackage{natbib}

\usepackage{amsfonts}       

\usepackage{bbm}          

\usepackage{amssymb,amsmath,amsthm, mathtools}
\usepackage{amscd}

\usepackage{caption, subcaption}

\usepackage{enumitem}

\usepackage{siunitx} 

\newtheorem{theorem}{Theorem}[section]
\newtheorem{definition}[theorem]{Definition}

\newtheorem{lemma*}{Lemma}

\makeatletter
\renewcommand*\env@matrix[1][*\c@MaxMatrixCols c]{%
  \hskip -\arraycolsep
  \let\@ifnextchar\new@ifnextchar
  \array{#1}}
\makeatother

\def\R{\mathbb{R}}

\def\0{\mathbf{0}}

\def\cX{\mathcal{X}}

\def\st{\medspace|\medspace}

\DeclareMathOperator*{\argmax}{argmax}

\DeclarePairedDelimiterX{\infdivx}[2]{(}{)}{%
  #1\;\delimsize\|\;#2%
}

\newcommand{\suchthat}{\;\ifnum\currentgrouptype=16 \middle\fi|\;}

\newenvironment{itemize*}%
  {\begin{itemize}%
  \vspace{-0.5cm}
    \setlength{\itemsep}{0pt}%
    \setlength{\parskip}{0pt}}%
  {\end{itemize}}
\newenvironment{enumerate*}%
{\begin{enumerate}
    \vspace{-0.5cm}
    \setlength{\itemsep}{0pt}%
    \setlength{\parskip}{0pt}}%
  {\end{enumerate}}

  {\list{}{\leftmargin=#1\rightmargin=#1}\item[]}%
  {\endlist}

\usepackage{float}

\icmltitlerunning{On the Robustness of Interpretability Methods}

\begin{document}

\twocolumn[
\icmltitle{On the Robustness of Interpretability Methods}


\icmlsetsymbol{equal}{*}

\begin{icmlauthorlist}
\icmlauthor{David Alvarez-Melis}{mit}
\icmlauthor{Tommi S. Jaakkola}{mit}
\end{icmlauthorlist}

\icmlaffiliation{mit}{MIT Computer Science and Artificial Intelligence Lab}

\icmlcorrespondingauthor{David Alvarez-Melis}{dalvmel@mit.edu}

\icmlkeywords{interpretability, transparency, robustness}

\vskip 0.3in
]


\printAffiliationsAndNotice{}


\begin{abstract}
	We argue that robustness of explanations---i.e., that similar inputs should give rise to similar explanations---is a key desideratum for interpretability. We introduce metrics to quantify robustness and demonstrate that current methods do not perform well according to these metrics. Finally, we propose ways that robustness can be enforced on existing interpretability approaches.
\end{abstract}

\section{Introduction}

Most current methods for interpreting complex models are \emph{prediction-based}, i.e., they operate at the level of a single individual input/prediction pair, producing an explanation for why the model predicted that output for that particular input. These methods and can be roughly divided into two categories: saliency and perturbation approaches. Methods in the former category use signal from gradients or output decomposition to infer salient features \citep{selvaraju2016grad, simonyan2013deep}. On the other hand, perturbation-based methods rely on querying the model around the prediction of interest to infer relevance of input features towards the output \cite{Ribeiro2016Why, AlvarezMelis2017Causal}.

Such saliency and perturbation methods offer many desirable properties: they have simple formulations, require little (or no) modification to the model being explained, and some of them are derived axiomatically \cite{Lundberg2017Unified}. Yet, these methods in their current form have important limitations too. For example, \citet{kindermans2017unreliability} showed that most saliency methods are not invariant under simple transformations of the input, and are very sensitive to the choice of reference point. 

Another, more general, argument commonly used against prediction-based interpretability methods is that `understanding' a complex model with a single point-wise explanation is perhaps too optimistic, if not naive. Indeed, the insight gained from a single attribution or saliency map might be too brittle, and lead to a false sense of understanding. One way to address this limitation would be to go beyond points and examine the behavior of the model in a neighborhood of the point of interest.

In light of this, here we argue that a crucial property that interpretability methods should satisfy to generate meaningful explanations is \emph{robustness} to local perturbations of the input. In its most intuitive form, such a requirement states that similar inputs should not lead to substantially different explanations. There are two main arguments for why robustness is a crucial property that interpretability methods should strive for. First, in order for an explanation to be valid around a point, it should remain roughly constant in its vicinity, regardless of how it is expressed (e.g., as saliency, decision tree, or linear model). On the other hand, if we seek an explanation that can be applied in a predictive sense around the point of interest as described above, then robustness of the simplified model implies that it can be approximately used \emph{in lieu} of the true complex model, at least in a small neighborhood.

In this context, the purpose of this work is to investigate whether popular gradient and perturbation-based interpretability methods satisfy robustness. For this, we first formalize the intuitive notion of robustness that we seek in the next section. Then, in Section 3, we show how various popular interpretability methods fare with respect to these metrics in various experimental settings. Finally, in Section 4 we summarize our findings and discuss approaches to enforce robustness in interpretability methods.


\section{Robustness} 
\begin{figure}[t]
	\centering
	\includegraphics[width=\linewidth]{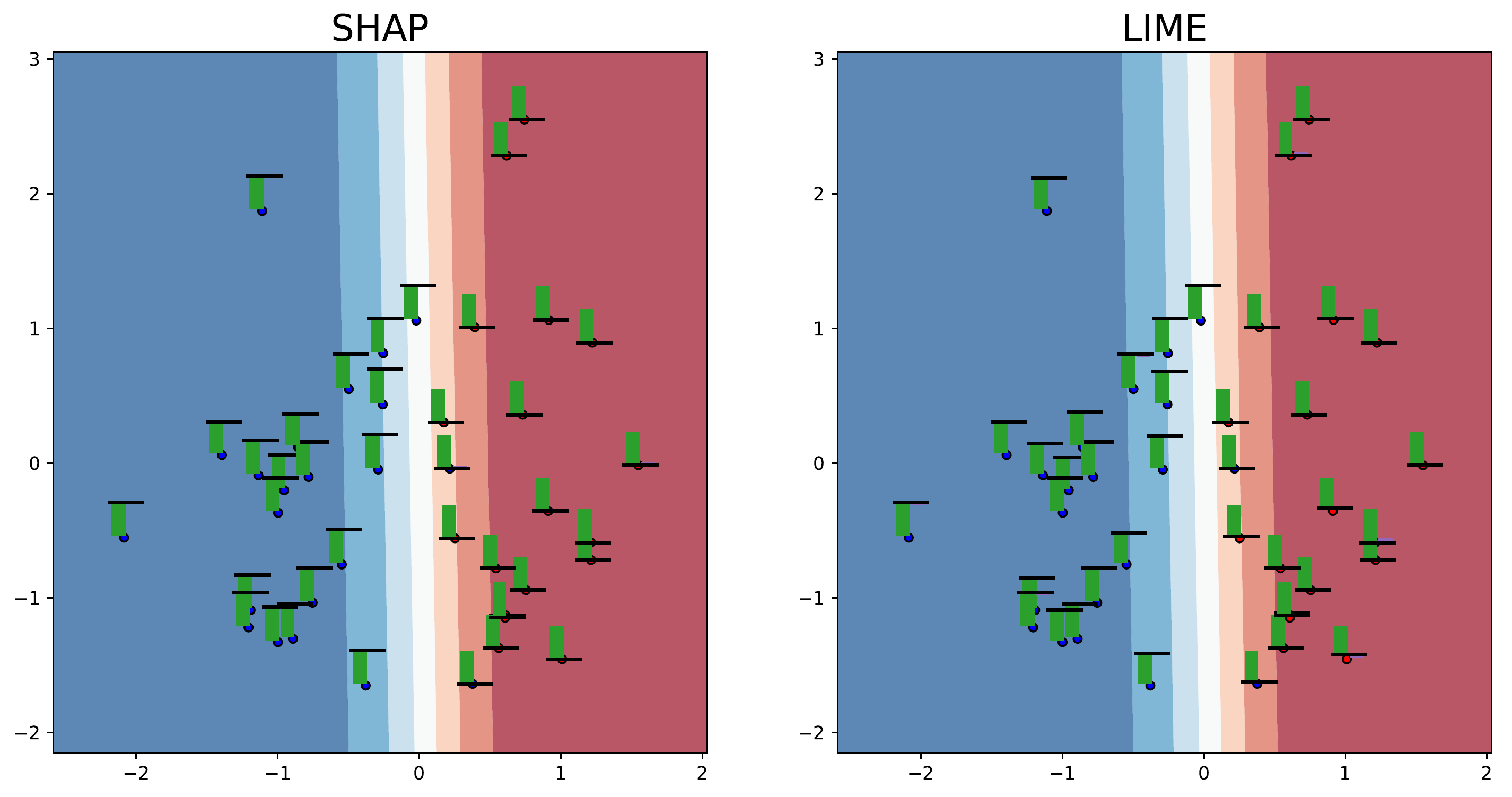}
	\includegraphics[width=\linewidth]{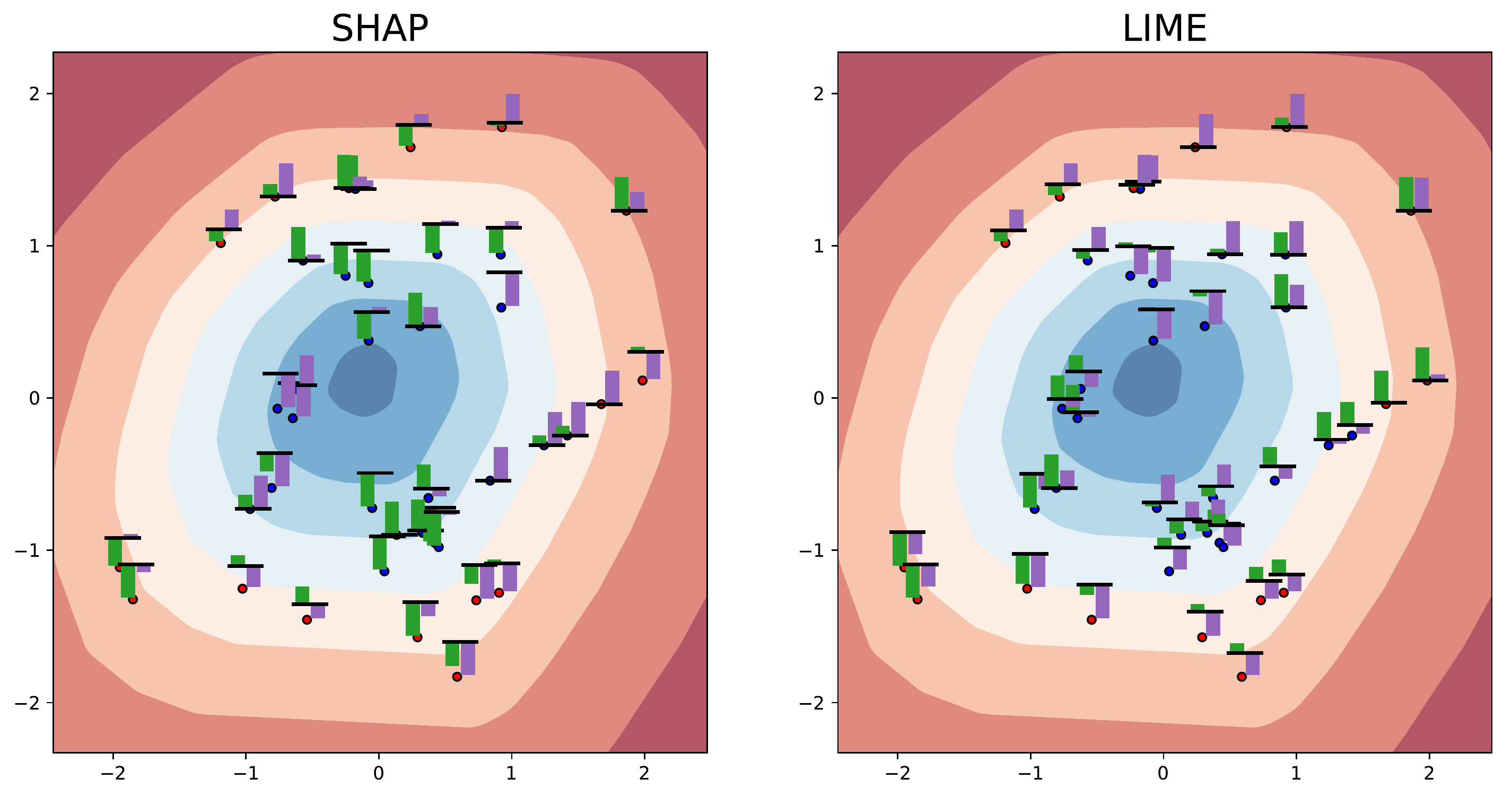}
	\caption{\textsc{Lime} and \textsc{Shap} explanations for two simple binary classifiers: a linear SVM (top row) and a two-layer neural network (bottom). The heatmaps depict the models' positive-class probability level sets, and the barchart inserts show the interpreters' explanations (attribution values for $x$ in green and $y$ in purple) for test point predictions. While both \textsc{Lime} and \textsc{Shap}'s explanations for the linear model are stable, for the non-linear model (bottom) they vary significantly within small neighborhoods.}\label{fig:synth_plots}
\end{figure}

The notion of robustness we seek concerns variations of a prediction's ``explanation'' with respect to changes in the input leading to that prediction. Intuitively, if the input being explained is modified slightly---subtly enough so as to not change the prediction of the model too much---then we would hope that the explanation provided by the interpretability method for that new input does not change much either. The first important takeaway from this work---and its main motivation---is that this is not the case for most current interpretability methods. Figure~\ref{fig:synth_plots} shows the explanations provided by two popular such perturbation-based methods, LIME \citep{Ribeiro2016Why} and SHAP \cite{Lundberg2017Unified}, for the predictions of two classifiers on a synthetic two-dimensional dataset. As expected, their predictions are fairly stable when explaining a linear SVM classifier (top row), but for a more complex model (a neural network classifier, shown in the bottom row), they yield explanations that vary considerably for some neighboring inputs, and are often inconsistent with each other. 

The instability portrayed in Figure~\ref{fig:synth_plots} is the phenomenon we seek to investigate. Visual inspection of attributions, although illustrative, is subjective and infeasible for higher-dimensional inputs. To conclusively gauge this (lack of) robustness, we need objective tools to quantify it. Calculus puts multiple notions of function stability at our disposal, among which is Lipschitz continuity, a parametric notion of stability that measures relative changes in the output with respect to the input. Note, however, that the usual definition on Lipschitz continuity is \emph{global}, i.e., it looks for largest relative deviations throughout the input space. In the context of interpretability, such a notion is not meaningful since there is no reason to expect explanation uniformity for very distant inputs. Instead, we are interested in a \emph{local} notion of stability, i.e., for neighboring inputs. Thus, we propose to rely on the point-wise, neighborhood-based \emph{local Lipschitz continuity}:\footnote{This notion has been also used for adversarial attacks on neural networks\citep{hein2017formal, weng2018evaluating}}
\begin{definition}\label{def:functional_local_lipschitz}
	$f: \cX \subseteq \R^n \rightarrow \R^m$ is \textbf{locally Lipschitz} if for every $x_0$ there exist $\delta >0 $ and $L \in \R$ such that $\| x - x_0 \| < \delta$ implies $\| f(x) - f(x_0)\| \leq L \|x - x_0\|$.
\end{definition}
As opposed to the (global) Lipschitz criterion, here both $\delta$ and $L$ depend on the anchor point $x_0$. Armed with this notion, we can quantify the robustness of an explanation model $f$ in terms of its constant $L$ in Definition~\ref{def:functional_local_lipschitz}. Naturally, this quantity is rarely known a-priori, and thus has to be estimated. A straightforward way to do so involves solving, for every point $x_i$ of interest, an optimization problem:
\begin{equation}\label{eq:eval_metric_continuous}
	\hat{L}(x_i) = \argmax_{ x_j \in B_{\epsilon}(x_i)} \frac{\| f(x_i) - f(x_j)\|_2}{\| x_i - x_j\|_2}
\end{equation}
where $N_{\epsilon}(x_i)$ is a ball of radius $\epsilon$ centered at $x_i$.\footnote{Naturally, optimizing over $l_{\infty}$ box constraints is much easier, and thus we take this approach in our experiments.} Computing this quantity is a challenging problem by itself. For our setting, most functions $f$ of interest (i.e., interpretability methods) are not end-to-end differentiable, so computing gradients with respect to inputs (e.g., for gradient ascent) is not possible. In addition, evaluations of $f$ are computationally expensive, so \eqref{eq:eval_metric_continuous} must be estimated with a restricted evaluation budget. There are various off-the-shelf methods for such black-box optimization, for instance Bayesian Optimization \citep[and references therein]{snoek2012practical}.

The \emph{continuous} notion of local stability described above might not be suitable for models with discrete inputs or those where adversarial perturbations are overly restrictive (e.g., when the true data manifold has regions of flatness in some dimensions). In such cases, we can instead define a (weaker) empirical notion of stability based on discrete, finite-sample neighborhoods, as implied by the examples in the test data of interest. Let $X=\{x_i\}_{i=1}^n$ denote a sample of input examples. Define, for every $x_i\in X$,  
\[ \mathcal{N}_{\epsilon}(x_i) = \{ x_j \in X \st \| x_i - x_j \| \leq \epsilon\} \]
The notion of interest is then
\begin{equation}\label{eq:eval_metric_discrete}
	\tilde{L}_X(x_i) = \argmax_{x_j \in \mathcal{N}_{\epsilon}(x_i)  \leq \epsilon} \frac{\| f(x_i) - f(x_j)\|_2}{\| x_i - x_j\|_2}
\end{equation}
Computation of this quantity, unlike \eqref{eq:eval_metric_continuous}, is trivial since it operates only over the (finite) test set $X$. 

Although both \eqref{eq:eval_metric_continuous} and \eqref{eq:eval_metric_discrete} are unitless quantities, there is no single ``ideal'' value that is universally desirable. Instead, what is \emph{reasonable} will depend on the application and goal of interpretability (see \S4). Here, we interpret these quantities relatively, comparing them across different methods.

\begin{figure}
	\centering
	\includegraphics[width=1.0\linewidth]{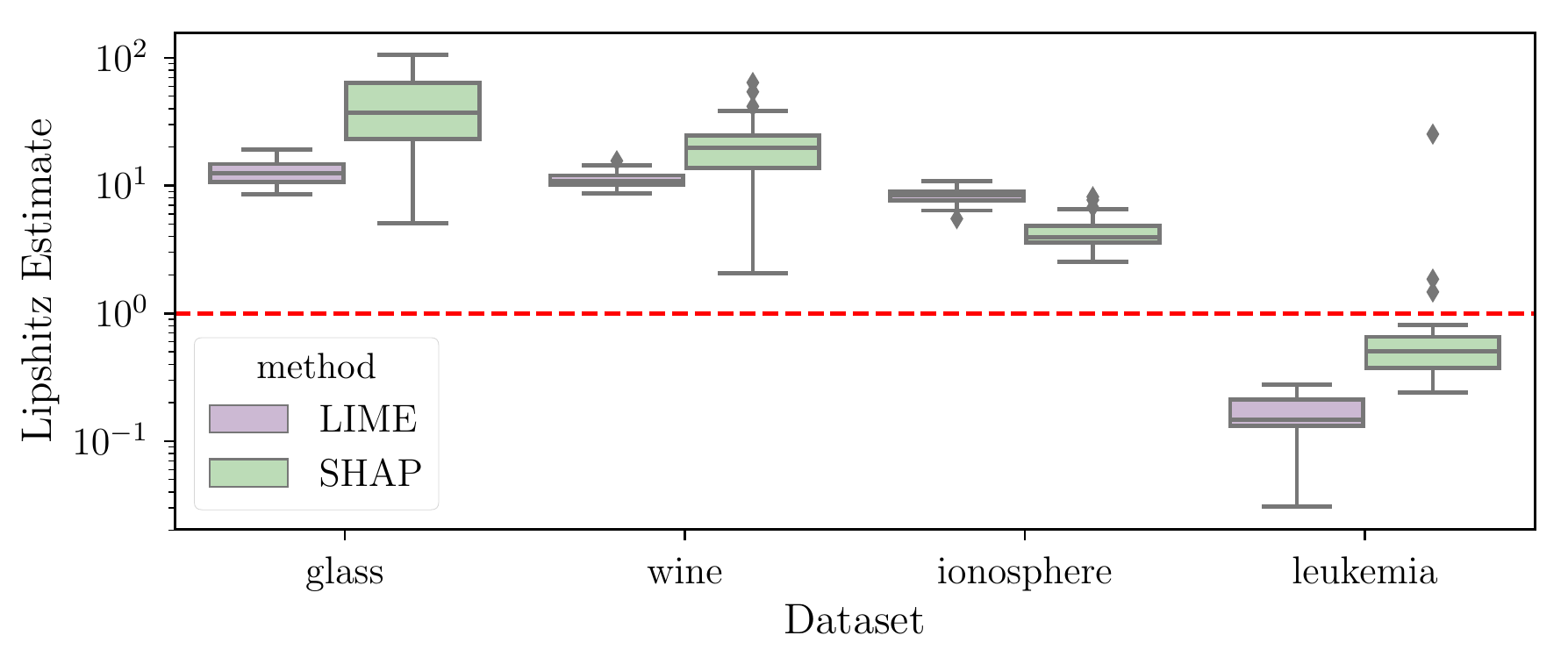}
	\caption{Local Lipschitz estimates \eqref{eq:eval_metric_continuous} computed on 100 test points on various UCI classification datasets.}\label{fig:uci_barchart}
\end{figure}
\section{Experiments} 
\label{sec:results}
\subsection{Methods and Datasets} 
\label{sub:model_and_datasets}

In addition to the aforementioned \textsc{Lime} and \textsc{Shap}, we compare the following interpretability methods:
\begin{itemize}[noitemsep,wide=0pt, leftmargin=\dimexpr\labelwidth + 2\labelsep\relax]
	\vspace{-0.3cm}
	\item \textsc{Saliency} maps \citep{simonyan2013deep}.
	\item \textsc{Gradient*Input} \citep{shrikumar2016not}.
	\item \textsc{Int}egrated \textsc{Grad}ients \citep{sundararajan2017axiomatic}.
	\item $\epsilon$-Layerwise Relevance Propagation \citep{Bach2015Pixel-wise}.
	\item \textsc{Occlusion} sensitivity \citep{zeiler2014visualizing}.
	\vspace{-0.3cm}
\end{itemize}
We used author implementations of \textsc{Lime} and \textsc{Shap} and the \texttt{DeepExplain}\footnote{\url{github.com/marcoancona/DeepExplain}} toolbox for the rest. All these methods return attribution arrays, which we treat as the vector-valued $f(x)$ in \eqref{eq:eval_metric_continuous} and \eqref{eq:eval_metric_discrete}. We compute the latter using Bayesian optimization with the \texttt{skopt}\footnote{\url{scikit-optimize.github.io}} toolbox, using a budget of $200$ function calls (only $40$ for \textsc{Lime}/$\textsc{Shap}$ due to higher compute time). We use $\epsilon = 0.1$ in \eqref{eq:eval_metric_continuous} and \eqref{eq:eval_metric_discrete}.

We test these methods on various dataset/prediction model settings. First, we experiment with explaining black-box classifiers on standard machine learning datasets from the UCI repository \cite{lichman2013uci} and the \textsc{Compas} dataset. Then, we consider two image-processing tasks: explaining the predictions of a convolutional neural network (CNN) classifier on the \textsc{Mnist} dataset \citep{LeCun1998} (Section 3.3) and a ResNet classifier \cite{He2016} on natural images from the \textsc{ImageNet} dataset (Section 3.4).

\begin{figure}
	\centering
    \begin{subfigure}[t]{0.498\linewidth}
        \centering
		\includegraphics[width=\linewidth]{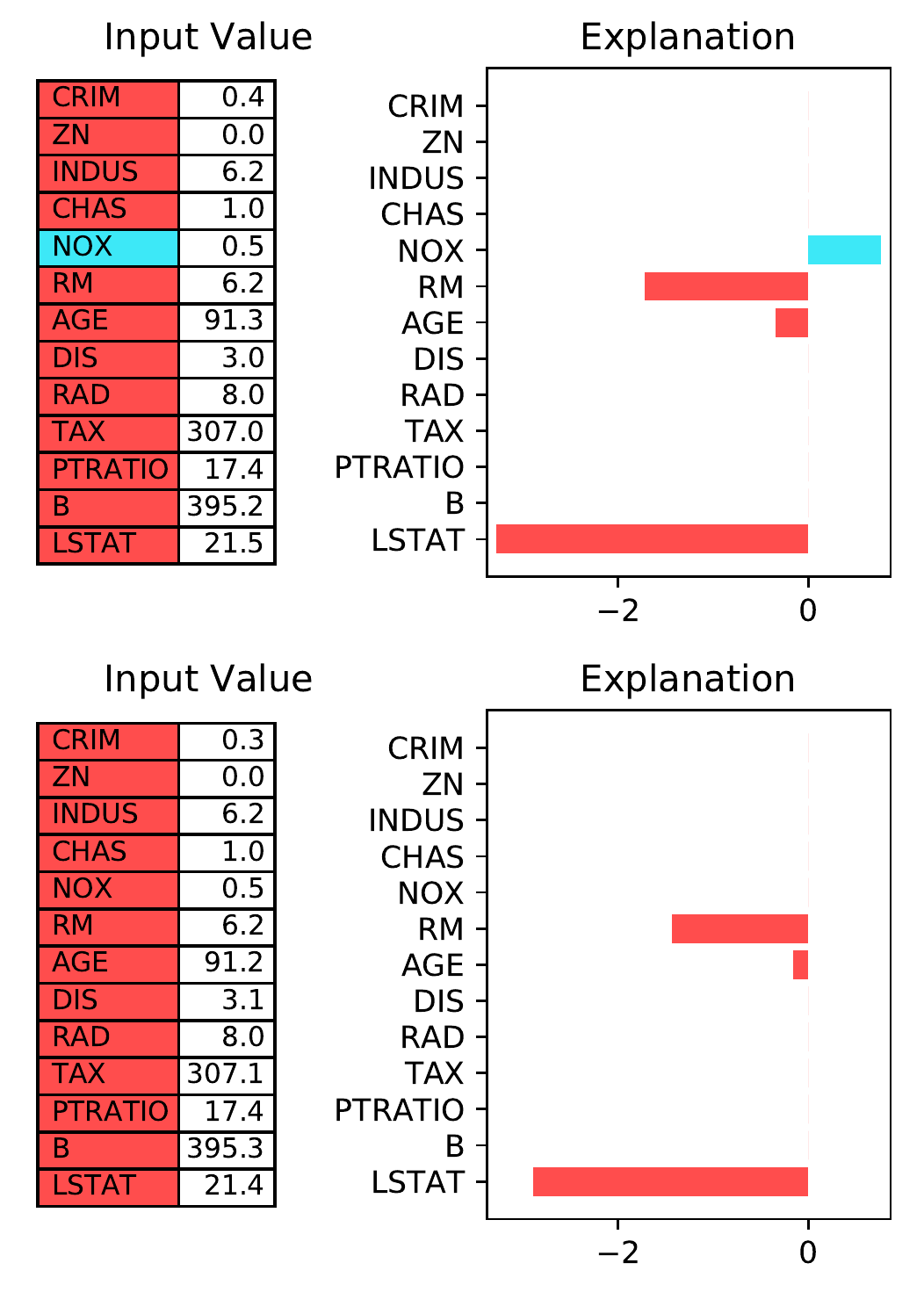}%
        \caption{\textsc{Shap} ($L=6.98$)}
    \end{subfigure}%
    ~ 
    \begin{subfigure}[t]{0.498\linewidth}
        \centering
		\includegraphics[width=\linewidth]{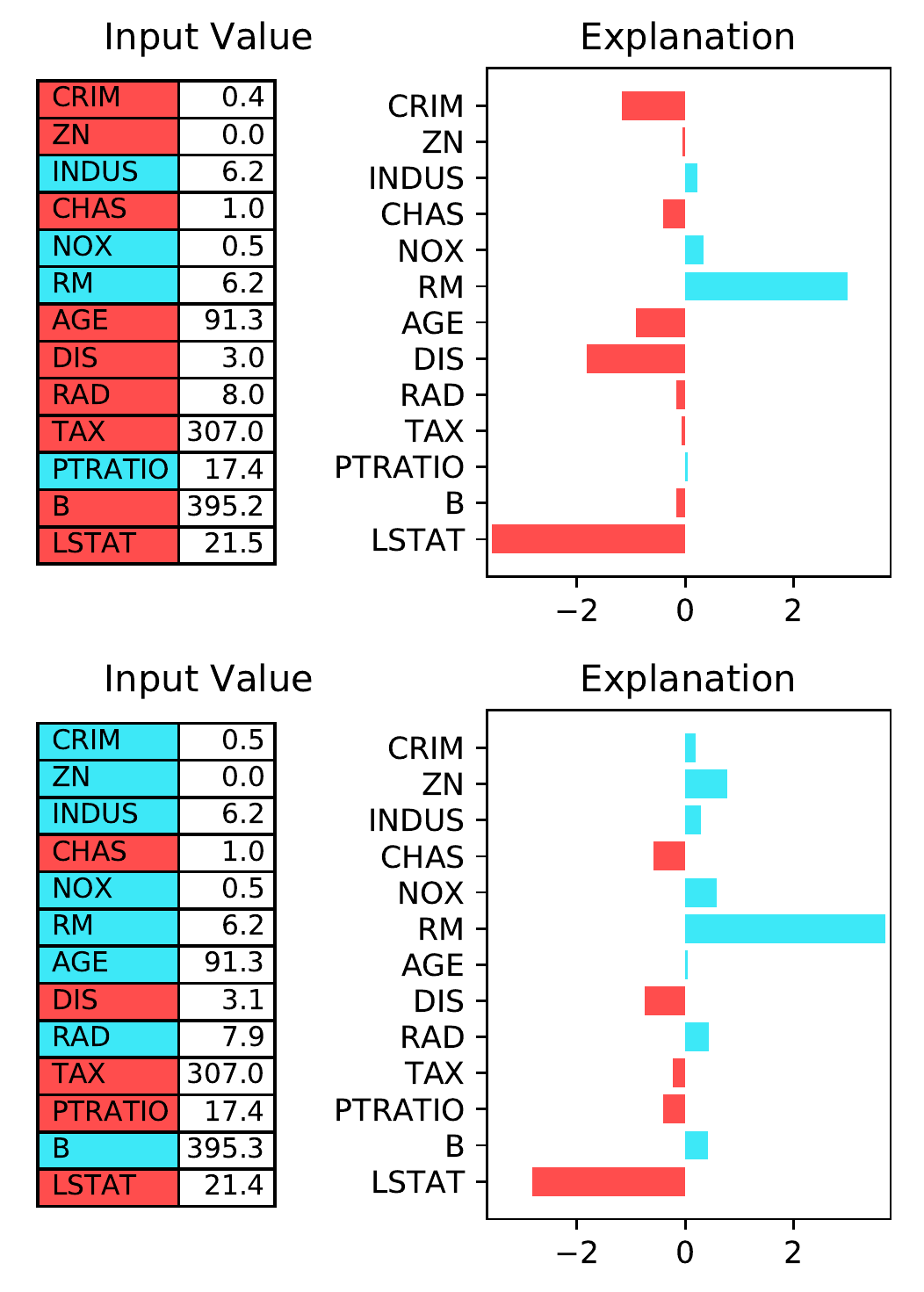}%
        \caption{\textsc{Lime} ($L=18.17$)}
    \end{subfigure}%
	\caption{\textbf{Top}: example $x_i$ from the \textsc{Boston} dataset and its \emph{explanations} (attributions). \textbf{Bottom}: explanations for the maximizer of the Lipschitz estimate $L(x_i)$ as per \eqref{eq:eval_metric_continuous}.}\label{fig:boston}
\end{figure}

\subsection{Benchmark Classification and Regression Datasets} 
\label{sub:uci_datasets}

In our first set of experiments, we evaluate the robustness of black-box interpretability methods (i.e., only \textsc{Lime} and \textsc{Shap} since all other methods considered require access to gradients or activations). For each dataset, we follow the same pipeline: (i) train a random forest classifier (or regressor) on the training data, (ii) randomly sample 200 points from the test set, (iii) use the interpretability methods to explain the predictions of the black-box model on them, and (iv) compute local robustness for each of these points by using \eqref{eq:eval_metric_continuous}. The aggregated results are shown in Figure~\ref{fig:uci_barchart}. 

It is illustrative to compare the explanations provided by each method for the model's prediction for some point $x_i$ and its adversarially chosen worst-case deviation, i.e., the  $x_j$ maximizing \eqref{eq:eval_metric_continuous} for that $x_i$. As an example, the examples from the \textsc{Boston} dataset shown in Figure~\ref{fig:boston} are extremely close but lead to considerably different explanations. 

\begin{figure}
	\centering
    \begin{subfigure}[t]{0.498\linewidth}
        \centering
		\includegraphics[width=\linewidth]{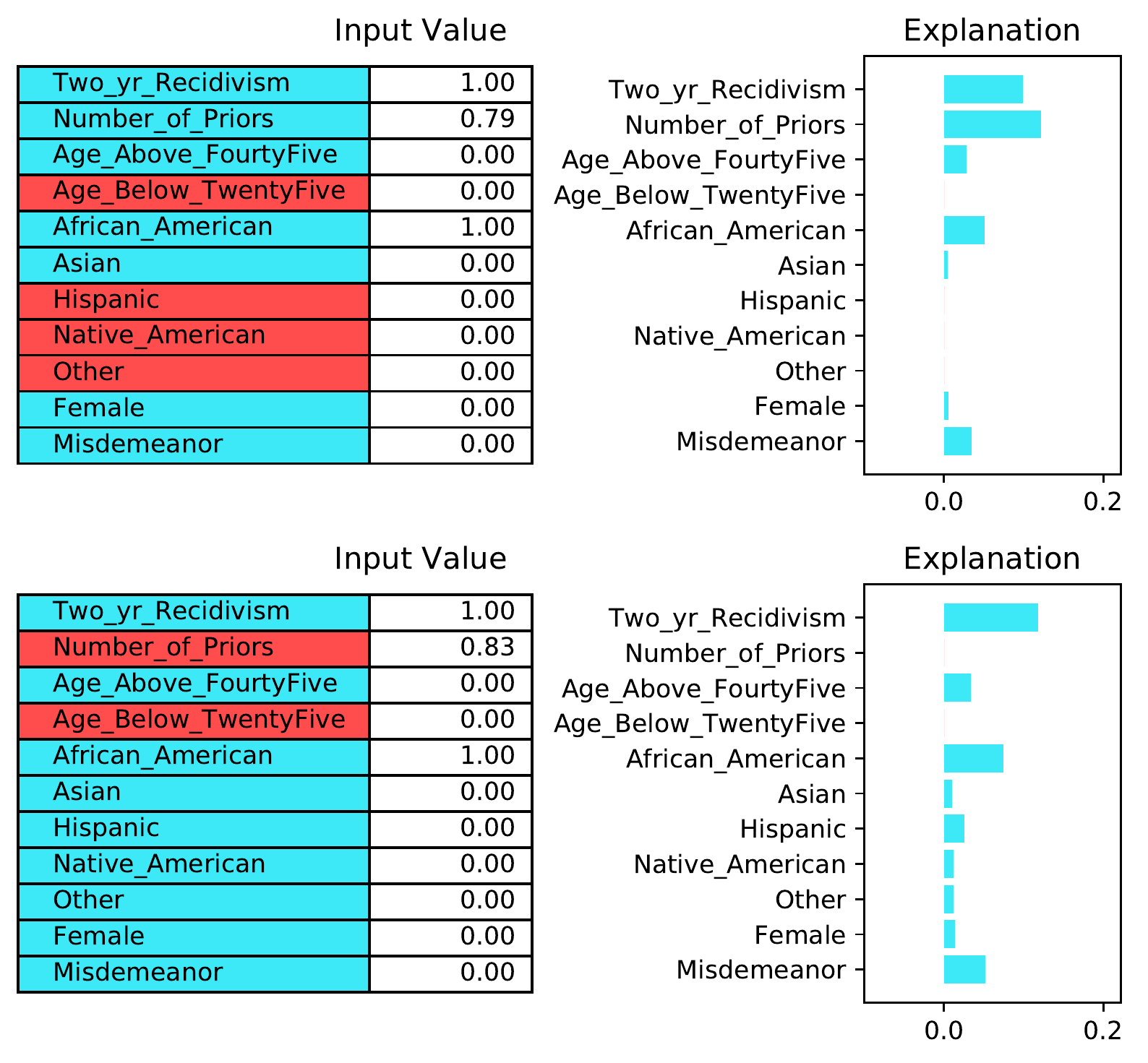}%
        \caption{\textsc{Shap} ($L=3.63$)}
    \end{subfigure}%
    ~ 
    \begin{subfigure}[t]{0.498\linewidth}
        \centering
		\includegraphics[width=\linewidth]{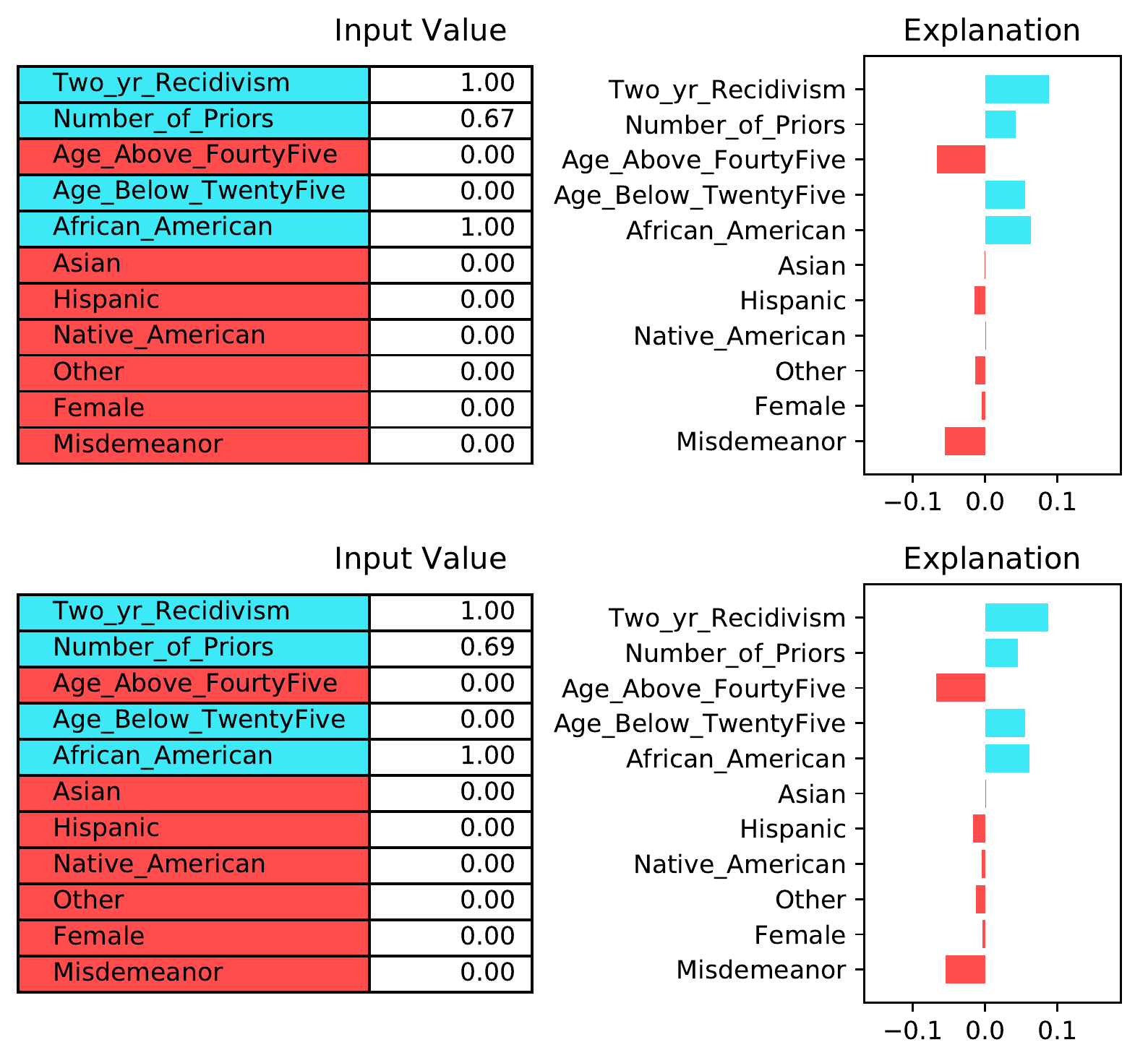}%
        \caption{\textsc{Lime} ($L=0.67$)}
    \end{subfigure}%
	\caption{Robustness upon explaining a classifier on the \textsc{Compas} dataset. The two rows correspond to the pair maximizing $\tilde{L}_X$ \eqref{eq:eval_metric_discrete} over the entire test fold, with $\epsilon=0.1$. }\label{fig:compas_argmax}
\end{figure}

The \textsc{Compas} dataset consists of categorical variables, and thus continuous perturbations are not very meaningful, as discussed in Section 2. Therefore, in this case we estimate robustness using the discrete, sample-based Lipschitz criterion \eqref{eq:eval_metric_discrete}, where we take the test set ($\sim600$ examples) as the reference sample. We use logistic regression as the classifier. In Figure~\ref{fig:compas_argmax} we show explanations for the pair of points with the largest (discrete) Lipschitz value.


\begin{figure*}
	\centering
	\includegraphics[width=0.98\linewidth]{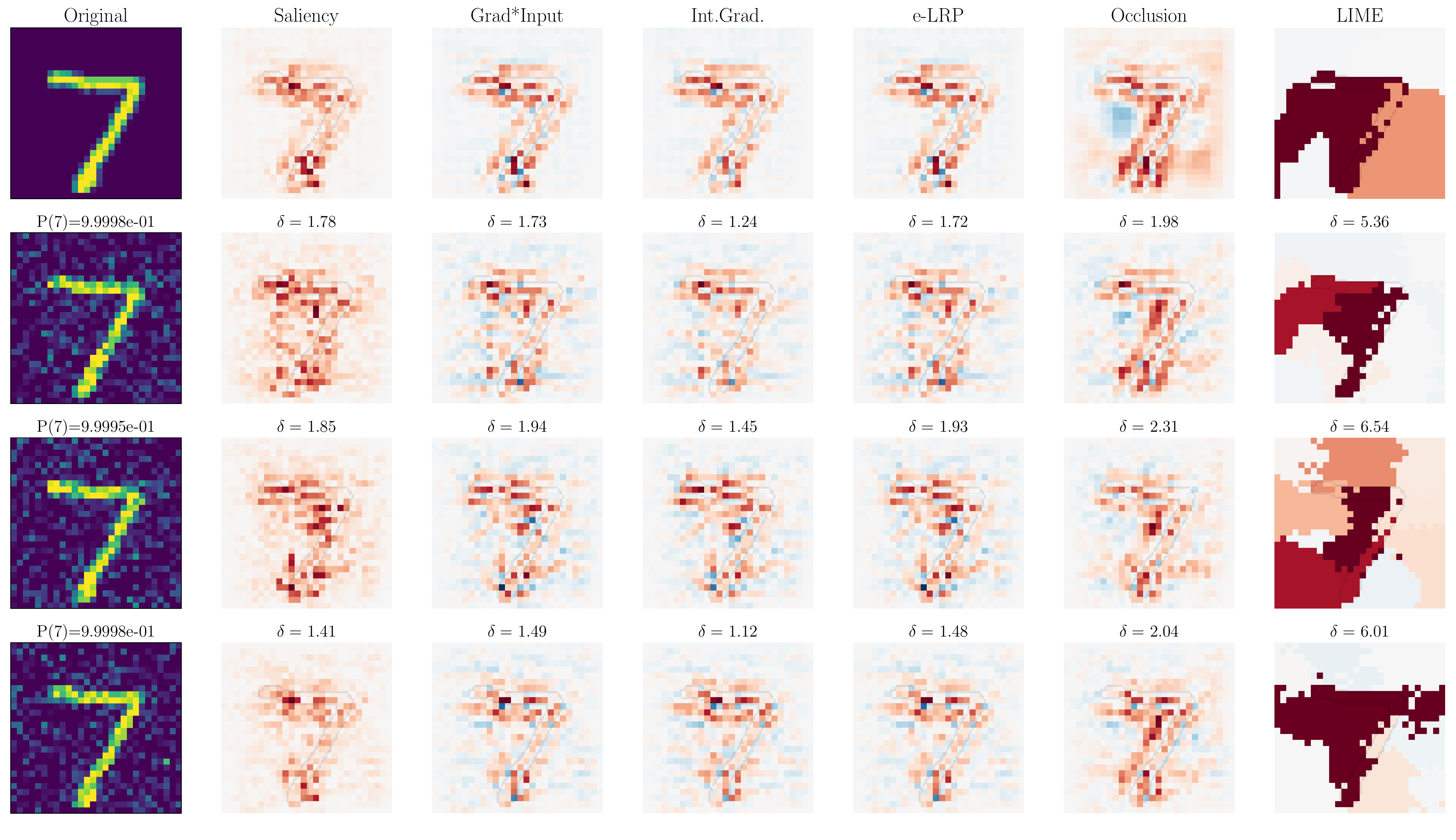}
	\caption{Explanations of a CNN model prediction's on a example \textsc{Mnist} digit (top row) and three versions with Gaussian noise added to it. The perturbed input digits are labeled with the probability assigned to the predicted class by the classifier. Here $\delta$ is the ratio $\| f(x) - f(x')\|_2/\| x- x'\|_2$ for the perturbed $x'$, which are not adversarially chosen as in \eqref{eq:eval_metric_continuous}.}\label{fig:gaussian_noise}
\end{figure*}

\begin{figure}[H]
	\centering
	\includegraphics[width=1\linewidth]{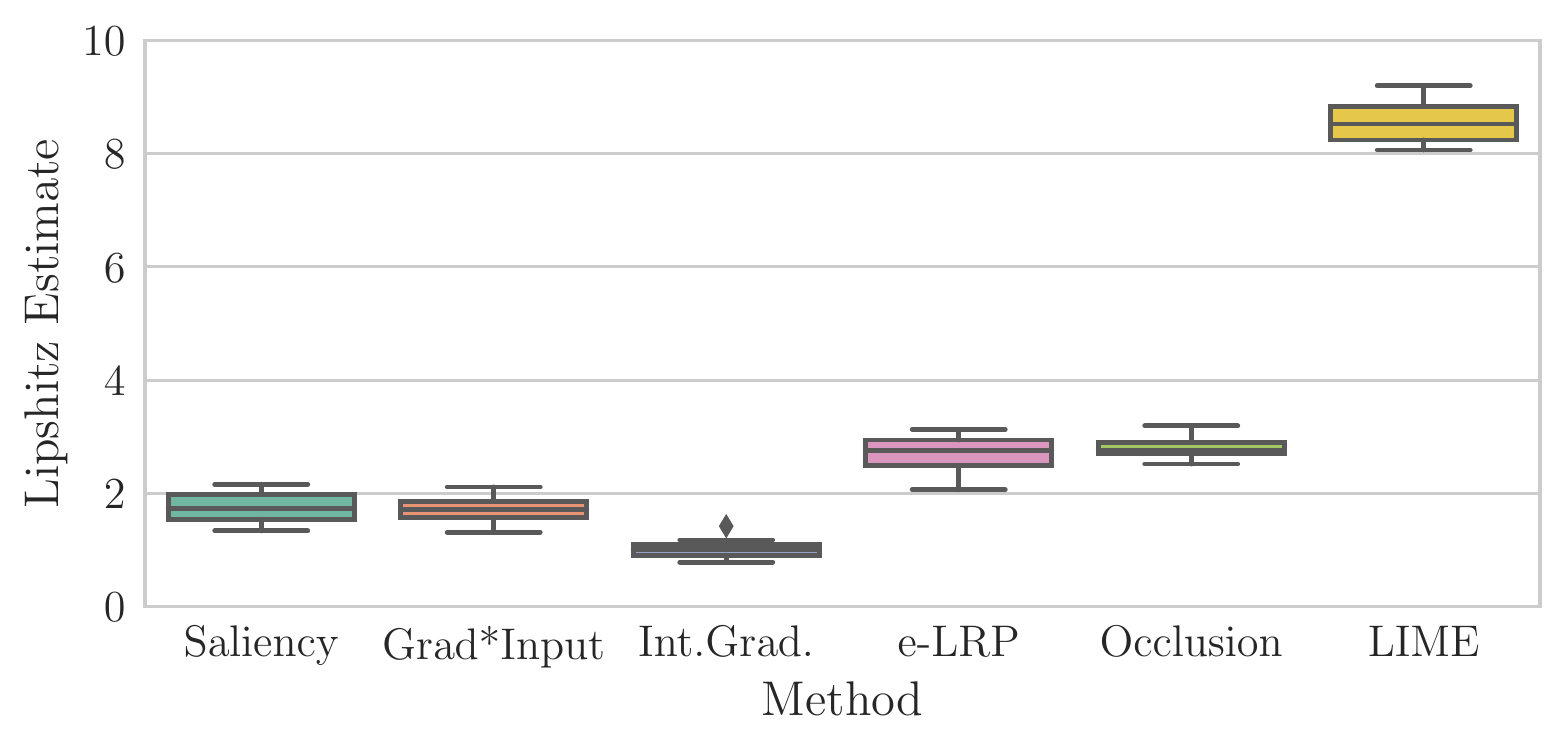}
	\caption{Local Lipschitz estimates computed according to \eqref{eq:eval_metric_continuous} on 100 test points on \textsc{Mnist} explanations.}\label{fig:mnist_boxplot}
\end{figure}

\vspace{-1cm}

\subsection{Explaining Digit Predictions} 
\label{sub:explaining_digit_predictions}

We first investigate the sensitivity of the interpretability methods in the presence of noise when explaining predictions of the digit classifier CNN trained on \textsc{Mnist}. For this, we take a test example digit and generate local perturbations by adding Gaussian noise to it. Figure~\ref{fig:gaussian_noise} shows the explanations provided by the various interpreters for the original input (top row) and three perturbations. Even though the classifier's predicted class probability barely changes as a consequence of these perturbations, the interpreter's explanations vary considerably, in some cases dramatically (\textsc{Lime}, \textsc{Occlusion}). 

\begin{figure*}
    \centering
    \begin{subfigure}[t]{0.33\textwidth}
        \centering
		\includegraphics[scale=0.4]{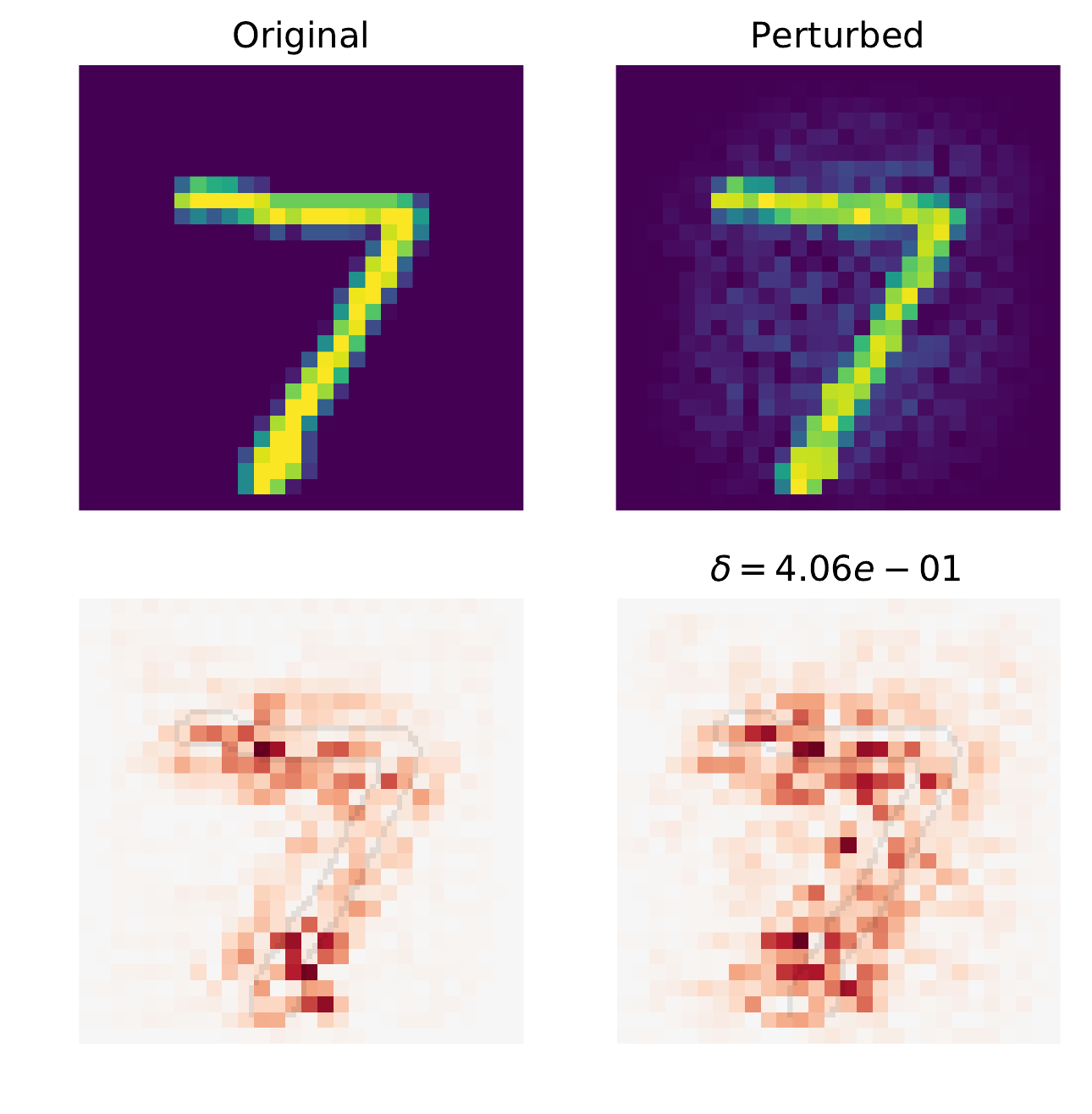}
        \caption{\textsc{Saliency}}
    \end{subfigure}%
    ~ 
    \begin{subfigure}[t]{0.33\textwidth}
        \centering
		\includegraphics[scale=0.4]{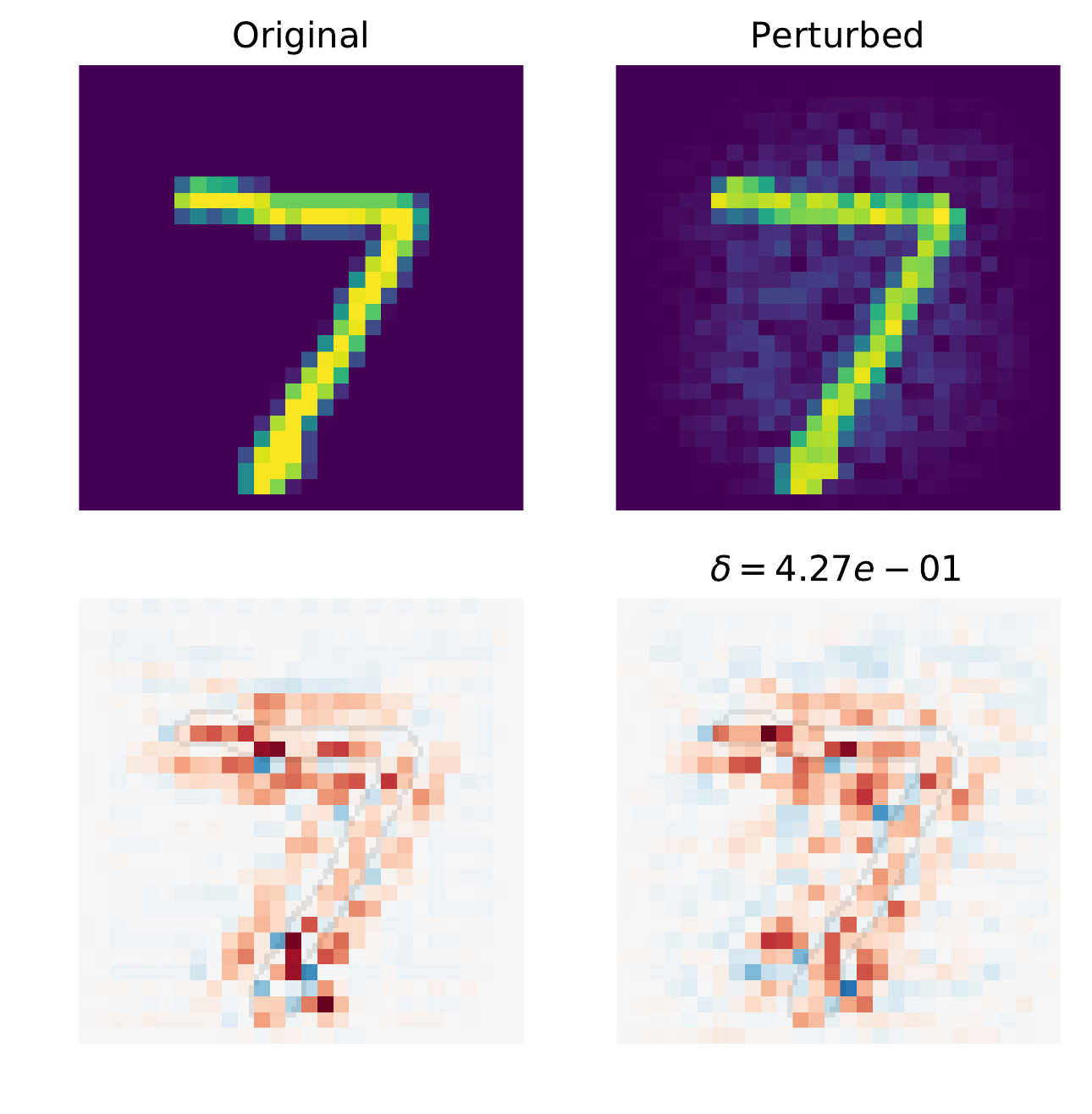}
        \caption{\textsc{e-LRP}}
    \end{subfigure}%
    ~ 
    \begin{subfigure}[t]{0.33\textwidth}
        \centering
		\includegraphics[scale=0.4]{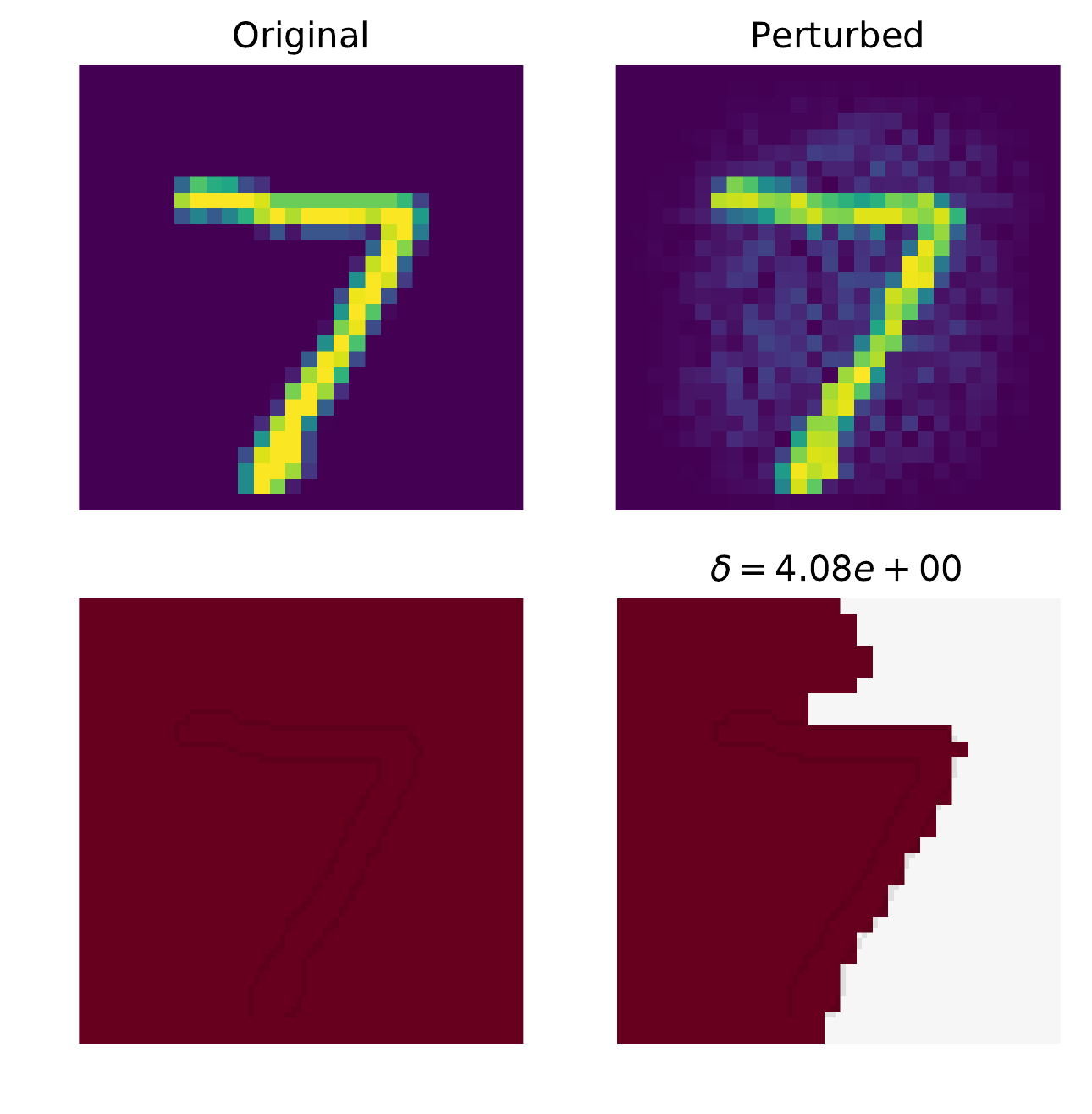}
        \caption{\textsc{Lime}}
    \end{subfigure}			
    \caption{True \textsc{Mnist} digits and their Lipschitz-maximizing perturbations with corresponding explanations.}\label{fig:mnist_argmaxes}
\end{figure*}

Again, we compute dataset-level robustness by repeating this procedure for multiple sample points in the test dataset (Figure~\ref{fig:mnist_boxplot}). In addition, we show in Figure~\ref{fig:mnist_argmaxes} the worst-case perturbations found through this procedure for a particular input. All methods are significantly affected by these minor perturbations, most notably \textsc{Lime}, whose sparse super-pixel based explanations make it particularly sensitive to small perturbations in the input.

\begin{figure}[t!]
	\centering
	\includegraphics[width=0.92\linewidth]{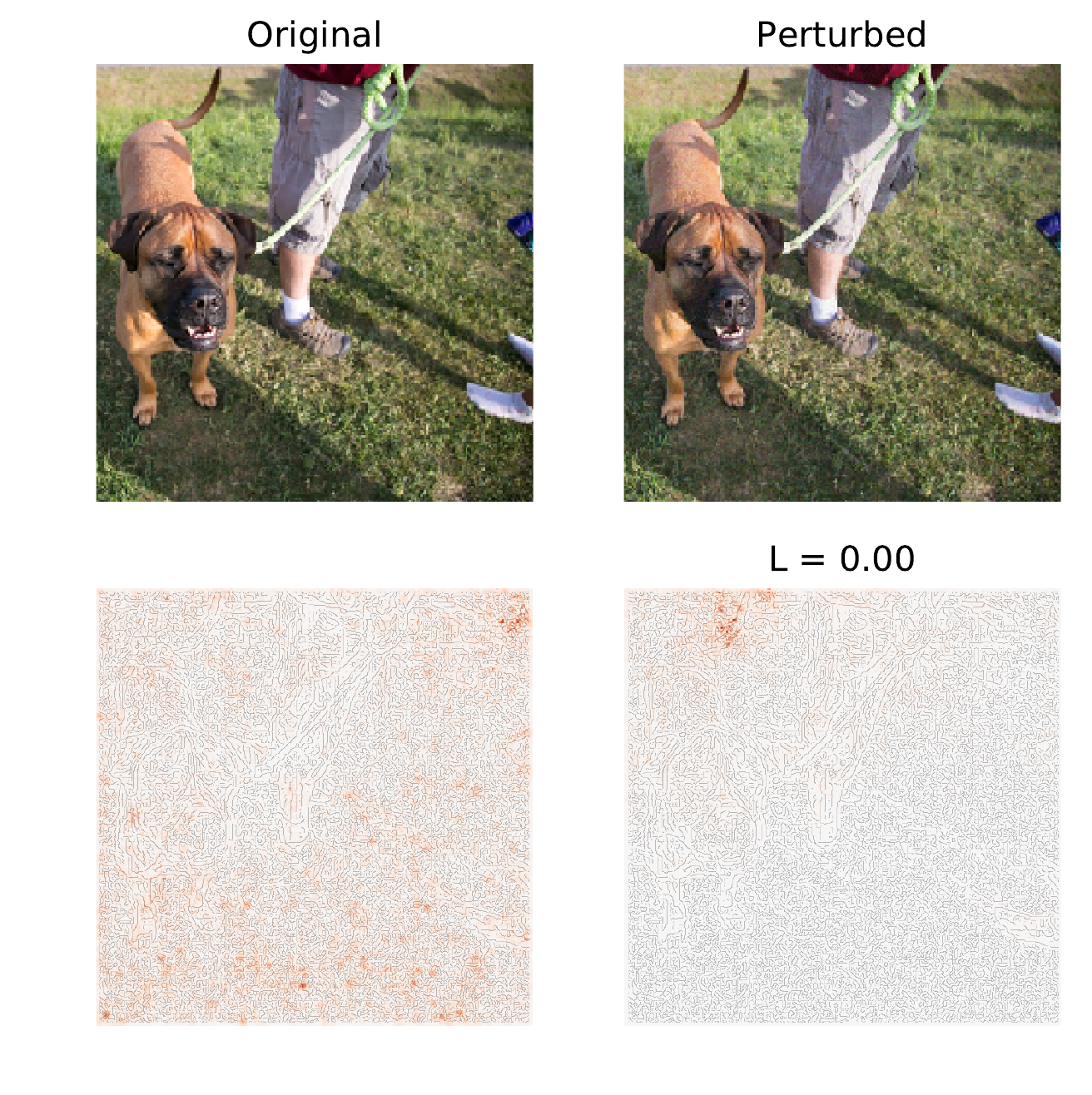}
	\caption{\textsc{Saliency} explanations for \textsc{ResNet} model prediction, and its Lipschitz-maximizing perturbation.}\label{fig:images}
\end{figure}

\subsection{Explaining Image Classification} 
\label{sub:explaining_image_classification}
We finalize by evaluating the robustness of the interpretability methods in the context of natural image classification. Now, we use various interpretability methods to explain a ResNet classifier trained on natural images at $224\times224$ pixel resolution. The size of these images makes it prohibitive to compute \eqref{eq:eval_metric_continuous} repeatedly to estimate dataset-level statistics, so we compute it only for a few images. Here, we show in Figure~\ref{fig:images} as an example the perturbed input maximizing the quantity \eqref{eq:eval_metric_continuous} for \textsc{Saliency}. The perturbed version of the image is mostly indistinguishable from the original input to the human eye, and the model predicts the same class (\texttt{bull\_mastiff}) in both cases with almost identical probabilities ($0.7308$ vs $0.7307$), yet the explanations are remarkably different.

\vspace{-0.3cm}


\section{Discussion} 
\label{sec:discussion}

In this work we set to investigate whether current popular interpretability frameworks are robust to small modifications of the input. Our experiments show that, for the most part, they are not, but that model-agnostic perturbation-based methods are (unsurprisingly) more prone to instability than their gradient-based counterparts. 

Here we focused on small perturbations that have minimal (or no) effect on the underlying model's predictions, yet have significant effects on the explanations given be the interpreters meant to explain them. Yet, a natural question is whether we should expect interpretability methods to be robust when the model being explained is itself not robust. As a concrete example, consider an image classification model that places importance on both salient aspects of the input---i.e., those actually related to the ground-truth class--- \emph{and} on background noise. Suppose, in addition, that those artifacts are not uniformly relevant for different inputs, while the `salient' aspects are. Should the explanation include the noisy pixels? 

While there in probably no absolute answer to this question, some use cases of interpretability allow for more definite statements. If the purpose of the explanation is to get a exact traceback of outputs to inputs (e.g., for debugging the model), then it is probably reasonable to have a broad definition of ``influence'', including such artifacts. If, on the other hand, the goal of interpretability is to gain understanding on \emph{both the predictor and the underlying phenomenon it is modeling}, then it is imperative the explanations focus on the stable relevant aspects of the input (e.g., those which are consistently used by the model in local neighborhoods), while ignoring unstable aspects. In this case, not only is it reasonable to expect the explanation method to be as robust as the underlying model, but rather, it is perhaps necessary to require it to be even more so.

A natural follow-up question is how to enforce such robustness into current interpretability methods, or how to design new ones that are robust \emph{by construction}. A slight generalization of criterion \eqref{eq:eval_metric_continuous} can be used to train interpretable neural networks with robust explanations \citep{AlvarezMelis2018Towards}. Alternatively, various techniques that share similar intuitive motivation with our framework have been proposed in the context of adversarial training of neural networks (e.g., \cite{kolter2017provable, raghunathan2018certified}) which could inspire approaches for interpretability robustness.  Additional notions of robustness found in that literature would make for interesting complementary evaluation metrics to the one proposed here.



\bibliography{RobustInterpet.bib}
\bibliographystyle{icml2018}

\end{document}